\documentclass{article}
\usepackage{spconf,amsmath,graphicx}
\usepackage{xcolor,graphicx,amssymb,multirow,multicol,booktabs}
\usepackage[usestackEOL]{stackengine}

\def\L{{\cal L}}
\ninept

\title{Contrastive Semi-supervised Learning for ASR}
\name{Alex Xiao, Christian Fuegen, Abdelrahman Mohamed}
\address{Facebook AI}
\begin{document}
%
\maketitle
\begin{abstract}
Pseudo-labeling is the most adopted method for pre-training automatic speech recognition (ASR) models.  However, its performance suffers from the supervised teacher model's degrading quality in low-resource setups and under domain transfer.
Inspired by the successes of contrastive representation learning for computer vision and speech applications, and more recently for supervised learning of visual objects, we propose Contrastive Semi-supervised Learning (CSL). CSL eschews directly predicting teacher-generated pseudo-labels in favor of utilizing them to select positive and negative examples. In the challenging task of transcribing public social media videos, using CSL reduces the WER by 8\% compared to the standard Cross-Entropy pseudo-labeling (CE-PL) when 10hr of supervised data is used to annotate 75,000hr of videos. The WER reduction jumps to 19\% under the ultra low-resource condition of using 1hr labels for teacher supervision. CSL generalizes much better in out-of-domain conditions, showing up to 17\% WER reduction compared to the best CE-PL pre-trained model.
\end{abstract}

\section{Introduction}
\label{sec:intro}
Driven by the practical need to deploy speech and audio models to new languages and domains, there has been a surge in research work utilizing self-, semi-, and weakly-supervised approaches for speech and audio representation learning achieving impressive results on a wide variety of tasks. 

The contrastive loss, which has been utilized extensively in self-supervised computer vision and speech applications~\cite{he2020momentum, cpc, w2v, w2v2}, centers around distinguishing one or more positive examples from negative ones given an anchor sample. It has also been utilized for supervised learning of visual objects~\cite{khosla2020supervised}.

Pseudo-labeling is the most adopted approach of semi-supervised learning in speech recognition since the late 1990s and more recently for modern end-to-end neural approaches~\cite{Kemp1999, Zavaliagkos_98, ma_bbn_06, wessel05, Bhuvana_05, wang_ICASSP07, singh2020large, gab_19, kahn2019self,park2020improved, tara_sslearning}. A 'teacher' model uses a small amount of supervised data for training, then transcribes a larger volume of unlabeled audio data. A 'student' model then utilizes the teacher-generated pseudo-labels for its training. Depending on the downstream task, the student model can be smaller~\cite{NIPS2014_5484} or larger~\cite{park2020improved} than the teacher model. A critical ingredient of pseudo-labeling is filtering the teacher-generated data by removing implausible, low-quality hypotheses through confidence filtering~\cite{hank2013}, which reduces the possibility of the student model copying the teacher's mistakes. The whole pseudo-labeling process stalls when the initial labeled data is not large enough to train a reliable teacher model.



In this paper, we introduce a Contrastive Semi-supervised Learning (CSL) approach that benefits from a contrastive loss for improving the stability of learned speech representations. Precisely, we use a contrastive loss to replace the cross-entropy loss in the standard pseudo-labeling pre-training (CE-PL). Contextual representations of audio segments are optimized to get closer to others with similar pseudo-labels (positive examples) and further away from ones that are different (negative examples). Such soft constraint during learning allows the CSL to be more resilient to teacher labeling noise.
To demonstrate its resilience to pseudo-labeling noise, we apply CSL pre-training in a low-resource setup with only 10hr of labeled data, where it reduces WER by 8\% compared to the standard cross-entropy pseudo-labeling (CE-PL). This WER reduction increase to 19\% with a teacher trained only on 1hr of labels and 17\% for out-of-domain conditions.



\section{Related Work}

Our paper builds upon previous research work in ASR pre-training using semi-supervised and self-supervised techniques. Pseudo-labeling, a semi-supervised method, where a supervised 'teacher' model is used to label a large volume of unlabeled data, is a widely adopted pre-training method for ASR. Similar approaches with slight variations exist in the literature as teacher-student training, knowledge distillation, and self-training. Since the 1990s, pseudo-labeling showed improvements to classical HMM/GMM systems as well as hybrid-NN systems~\cite{Kemp1999, Zavaliagkos_98, ma_bbn_06, Bhuvana_05, wessel05, wang_ICASSP07}, and also for end-to-end ASR approaches~\cite{singh2020large, gab_19, kahn2019self,park2020improved, tara_sslearning}. Our work inherits the same setup of utilizing a (small) set of supervised data for the teacher model learning similar to pseudo-labeling. However, we introduce a contrastive pre-training loss to deal with low-quality teacher pseudo-labels. 

Recently, self-supervised ASR pre-training achieved impressive performance, especially with minimal amounts of labeled data~\cite{cpc, w2v, effectiveness2019, APC, liu2020mockingjay, ling2020deep, w2v2, hubert}. Given unlabeled audio data, self-supervised methods rely primarily on pretext pre-training tasks that acquire their labels either from the input signal itself, e.g., predicting future frames~\cite{APC} or via unsupervised means, e.g., feature clustering~\cite{w2v2, hubert}. 
Many self-supervised methods boil down to contrasting positive and negative samples for a particular anchor sample. Depending on the downstream application in mind, positive samples could be near-by frames to the current time step or random samples for the same speaker. Negative samples could be further-away frames or samples for a different speaker~\cite{cpc, w2v2}. After pre-training, a supervised fine-tuning stage optimizes the model performance for one or more downstream tasks, e.g., ASR.

Our approach relies on pseudo-labels from a supervised teacher model for positive and negative example selection. This work draws inspiration from the recent application of contrastive learning for supervised learning of visual object classification~\cite{khosla2020supervised}.

\section{Method}
\subsection{Pseudo-Labeling}
The pseudo-labeling approach uses a supervised teacher ASR model to decode a large volume of unlabeled audio data $X$ and generate pseudo-labels $Y$. 
The model training proceeds by using these pairs in a supervised learning setup. Following a hybrid-NN training approach, we align the labels to the input feature vectors to get a sequence of frame-level pairs: ${(x_1, y_1), (x_2, y_2), (x_3, y_3), ...,(x_T, y_T)}$, where $y_i$ are represented as chenone units~\cite{le2019senones}. 

Let the model be composed of two functions $F_{enc}$ and $G_{pred}$. $F_{enc}(X_{1:T})$ encodes the full input sequence and returns a sequence of contextualized representations $Z_{1:T}$. The predictor network $G_{pred}(Z)$ produces a sequence of frame-level probability distributions over the chenone output units $p(\hat{Y}_{1:T}|X)$. In this paper, $G_{pred}$ is chosen to be a linear layer followed by a softmax function. 


Our training objective is the cross-entropy loss of predicting frame-level chenones:
  \vspace{-0.1cm}
 \begin{align*}
    CE &= -\frac{1}{\sum_{i=1}^B T_i} \sum_{i=1}^B \sum_{t=1}^{T_i} \log p(\hat{y}_{ti} = y_{ti}|X_i)
 \end{align*}
  \vspace{-0.1cm}
Where $B$ is the batch size, $T_i$ the length of utterance $i$, and $y_{ti}$ is the chenone target for utterance $i$ at frame $t$.

\subsection{Contrastive Semi-supervised Learning}
Contrastive losses, e.g., InfoNCE~\cite{cpc}, are at the heart of many self-supervised representation learning approaches, where two groups of samples, positive and negative, are selected for a specific anchor data point within a pretext task—the training phase proceeds by iteratively improving the model's ability to distinguish samples in each group. Negative samples are selected randomly from a mini-batch or a memory bank of previous experiences, while positive samples could be augmented versions of the anchor, nearby frames, or samples from the same speaker. Data points' designation as positive and negative samples determines the learned representation nature and performance on downstream tasks.

Our proposed Contrastive Semi-supervised Learning (CSL) approach synthesizes pseudo-labeling and self-supervised methods while solving some of their weaknesses. It bypasses the challenge of positive and negative sample selection of self-supervised methods by utilizing a supervised teacher. Using a large volume of unlabeled data and given the teacher's pseudo-labels, we select audio segments within one mini-batch as positive pairs if they share the same label and as negative pairs otherwise. Compared to standard pseudo-labeling methods, CSL is resilient to errors in teacher-generated targets since it utilizes relative distance between label classes as a learning signal.


More precisely, during the CSL pre-training, our model consists of two functions: The encoder $F_{enc}(X_{1:T}) = Z_{1:T}$ which encodes input audio into latent representations, and a projection network $M_{proj}(Z_{1:T})=H_{1:T}$ that maps encoder representations into a new space suitable for applying the contrastive loss. We normalize the projected features $H_{1:T}$ to unit length. A hybrid-NN supervised teacher generates aligned pseudo-labels $Y_{1:T}$ to guide the selection of positive and negative samples for the contrastive loss. 

Given the speech signal's local smoothness, each positive or negative sample is an audio segment with a specific label. Concretely, let's define a span $[i:j]$ as a segment if all its pseudo-labels $[y_i:y_j]$ are equal, and it is not contained in a larger span with the same pseudo-label. We then randomly sample a single time step $k$ to represent this segment with label $y_k$ in the loss function. Assuming $S$ segments in all utterances in the mini-batch, a total of $S$ representative latent vectors are selected $\{z_1,..,z_i,..,z_S\}$, then projected using $M_{proj}$ into unit length vectors $\{h_1,..,h_i,..,h_S\}$. Given the corresponding pseudo-labels $\{y_1,..,y_i,..,y_S\}$ for sampled vectors, our training objective is: 
 \vspace{-0.1cm}
 \begin{equation*}
     \L = \frac{1}{S} \sum_{i=1}^{S} \frac{1}{|P(i)|} \sum_{h_{p} \in P(i)} -\log \frac{\exp(h_i \cdot h_{p} / \tau)}{\sum_{h \in N(i) \cup \{h_p\}} \exp(h_i \cdot h / \tau)}
 \end{equation*}
$P(i)$ is the set of samples in the mini-batch with the same pseudo-label as $i$. The rest of the mini-batch samples belong to $N(i)$ as negative examples. $\tau$ is a tunable temperature hyperparameter. 


Following the InfoNCE loss, the CSL objective function adjusts the encoder $F_{enc}$ to bring its features of similar pseudo-labels closer and other negative samples further away. After the CSL pre-training, the prediction network $G_{pred}$ replaces the projection network $M_{proj}$ for supervised fine-tuning of the whole network, including the encoder network $F_{enc}$. 

In this paper, we apply frame-level cross-entropy fine-tuning, but other loss functions, e.g., the Connectionist Temporal Classification (CTC)~\cite{ctc} loss, can also be used. Our formulation is an adaptation of the previous work in computer vision~\cite{khosla2020supervised} which applies a contrastive loss in a supervised setup. The CSL pre-training offer many benefits:
\begin{itemize}
    \item Utilizing teacher pseudo-labels for selecting positive and negative samples, CSL is more stable than self-supervised pre-training methods, which are sensitive to the diversity and the criterion for choosing positive and negative samples. Moreover, CSL enables reliable sampling of positive examples within and across utterances in the mini-batch.  
    \item Unlike the standard pseudo-labeling pre-training methods, which forces the model to predict the teacher targets, CSL applies a softer constraint on learned representations through the contrastive loss. Such formulation improves robustness to noisy teacher pseudo-labels.
    \item Applying the contrastive loss over normalized representations emphasizes hard positives and negative examples, as shown analytically in~\cite{khosla2020supervised}, which enables the pre-trained model to generalize better under out-of-domain conditions.
\end{itemize}

\subsection{Label-Aware Batching (LAB)}
Large mini-batch sizes are critical for contrastive losses to ensure a diverse set of negative samples. For CSL, it also brings several positive realizations of speech sounds. Given the CSL loss formulation, we require at least two positive instances of each label class in the mini-batch. Some rare sounds might not form positive pairs for small randomly sampled mini-batches, leading to poor representations. To account for this problem, we propose Label-Aware Batching (LAB). LAB incrementally builds each mini-batch considering the number of segments representing label $k$ in the current partially-created mini-batch, $C(k)$. First, it samples a rare label $r$ with probability $(\frac{1}{C(r)})^\alpha / \sum_{i=1}^{n} (\frac{1}{C(i)})^\alpha$, and then samples without replacement two random utterances containing the class $r$ to add to the current mini-batch. LAB updates the counts $C(k)$ for all labels appearing in these two utterances and repeats until it reaches the maximum number of utterances in a mini-batch. We use $\alpha=2$ in our experiments. For stable learning, we aggregate gradients over multiple mini-batches to perform a single model update and scale the learning rate accordingly~\cite{goyal2017accurate}. 

\section{Experiments}

\subsection{Data}
We use two in-house de-identified data sources with no personally identifiable information (PII). The first source contains de-identified public five-minute videos on Facebook in British English and Italian. 
During training, we segment all videos to a maximum of 10 seconds. To simulate a low-resource condition, we use 10hr labeled data for both teacher training and the final fine-tuning stage. We also test a more extreme setup with merely 1hr worth of labels. We use 75,000hr of unlabeled videos for pre-training and evaluate performance on 14hr of human-labeled videos for each language. As an upper bound, we compare performance against a fully supervised setting with 650hr and 3,700hr labeled data for British English and Italian, respectively. We sample 23hr general English videos to test accent generalization. 
The second source contains recordings of crowd-sourced workers responding to artificial prompts with mobile devices. We collect two evaluation datasets to test out-of-domain generalization: 15hr of short message dictation and 13hr of long-form conversations of up to 144 second long. The language model (LM) for British English uses the 650hr transcripts plus an additional 13,000hr of general English video transcripts, while the Italian LM uses only the 3,700hr transcripts.

\subsection{Model Details}
$F_{enc}$ has a VGG front-end and 12 transformer blocks (dim=768, FFN=3072, 8 heads, stride 4) following~\cite{wang2020} (~90M parameters). For CSL pre-training, $M_{proj}$ is a single hidden layer network of size 1024 and 128 output dimension. Input and output representations of $M_{proj}$ are length normalized. 
Our input speech features are 80 dimensional, speed perturbed~\cite{kaldi_augment} Mel-scale log filterbank coefficients computed every 10ms over 25ms windows. We report the results of a single pass decoding using a 5-gram LM.

\subsection{Training Details}
For the pseudo-labeling baseline and CSL pre-training, the teacher model is a hybrid-NN ASR system trained on the initial 10hr of labeled data. Given the small number of supervised labels, we use 368 and 424 chenone outputs~\cite{le2019senones} for the British English and Italian systems, respectively. A supervised frame-level cross-entropy (CE) fine-tuning stage follows pre-training for all baseline and CSL models presented using the same amount of labels used for the respective teacher (either 10hr or 1hr depending on the experiment). 

We use the Adam optimizer~\cite{adam} with mixed-precision training~\cite{fp16} and gradient norm clipping at 10. We tune the learning rate (lr) for best performance on each task: lr=1e-4 for all supervised and pseudo-labeling baselines, lr=5e-5 for fine-tuning CSL pre-trained models. 
We apply 320k model update steps with a tri-stage lr schedule for pre-training: ~13k warm-up steps, and the remaining steps are split evenly between constant lr and linear decay. We use 16 GPUs for pre-training, with about 320sec of audio per GPU. We apply 60 epochs of updates on a single GPU with a 0.8 multiplicative decay every epoch after the 40th epoch for fine-tuning. Using the CSL loss introduced a 15\% increase in training time compared to baseline pseudo-labeling due to the pairwise similarity score computations, which can be further optimized by reducing the number of positive and negative examples used for frequent units. 

We apply two types of input feature masking: Double masking policy (LD)~\cite{spec_augment} and a Short Time Masking (STM) policy which follows LD in frequency masking and increases the number of time masks to 15 with lengths uniformly sampled between 16 and 32 timesteps.

\begin{table}[htb]
\centering
\setlength\tabcolsep{2.0pt}
 \begin{tabular}{cccc} 
\toprule
 & & \textbf{British English} & \textbf{Italian}\\ 
 \midrule\midrule
 \multicolumn{4}{c}{\emph{\textbf{Supervised Baseline}}} \\
 \midrule\midrule
 A1&Full supervised data & 23.1 (650hr) & 11.9 (3,700hr)  \\ 
 A2& 10hr of labels & 50.7 & 31.8  \\ 
 \midrule\midrule
 \multicolumn{4}{c}{\emph{\textbf{CE Pseudo-labeling (CE-PL) with 10hr of labels}}}\\
\midrule\midrule
 B1&Initial policy & 37.5& 19.6  \\ 
 B2& B1 + STM masking & 34.2 &  18.1  \\ 
 B3& B1 + C9 settings & 32.0 & 17.2  \\ 
 \midrule\midrule
\multicolumn{4}{c}{\emph{\textbf{Contrastive Semi-supervised Learning (CSL) with 10hr of labels}}} \\
\midrule\midrule
 C1&Initial policy & 36.4 & 19.1 \\ 
 C2& C1 + 420 seconds batch & 35.5 & 18.9 \\ 
 C3& C1 + 600 seconds batch & 34.8& 18.6 \\ 
 C4& C2 + 8 random negatives  & 35.9& 19.1  \\ 
 C5& C2 + 8 same uttt. negatives  & 35.3 &  18.7  \\ 
 C6& C5 + LAB  & 34.8& 18.3 \\ 
 C7& C5 + 4x grad accumulation & 34.1 & 18.1 \\ 
 C8& C5 + STM masking & 31.9 &  17.6 \\ 
 C9&  \textbf{C3+C5:C8}  & \textbf{29.4}& \textbf{16.0}  \\
\bottomrule
\end{tabular}
\caption{WER of supervised, pseudo-labeling, and CSL systems.}
\label{tab1}
\end{table}

\begin{table}
\centering
\setlength\tabcolsep{6.0pt}
 \begin{tabular}{cc} 
\toprule
 & \textbf{British English (WER)}  \\ 
 
 \midrule\midrule

 100\% CSL & 35.3 \\
50\% CSL + 50\% CE-PL & 36.6 \\
25\% CSL  + 75\% CE-PL & 36.8 \\
10\% CSL  + 90\% CE-PL  & 37 \\
\bottomrule
\end{tabular}
\caption{The effect of mixing the CSL and CE-PL losses.}
\label{tab2}
\vspace{-3mm}
\end{table}

\section{Results and Discussion}

\subsection{The main results}
Table \ref{tab1} presents the main results of our experiments. The best CE pseudo-labeling (CE-PL) pre-training strategy improves the respective supervised baseline by about 36\% and 46\% for both languages using 10hr of labels. CSL provides a relative improvement of 8\% and 7\% over the best CE-PL system bringing the overall WER closer to the fully supervised case, which utilizes orders of magnitude more labeled data. We observe performance degradation without normalizing the projected features. The $M_{proj}$ output dimension can be reduced to 128 without significant performance loss. Unlike the supervised case in~\cite{khosla2020supervised}, high values for the contrastive loss temperature ($\tau$=1) consistently provided the best performance since it reduces the harm of noisy teacher pseudo-labels. The proposed Short Time Masking (STM) policy improves both CE and CSL by a 6-9\% relative (B2, C8) compared to the baseline policy. Fine-tuning the whole network after pre-training offered better performance than freezing the $F_{enc}$ subnetwork.

\subsubsection{Effect of large and label-aware batching}
Increasing the per-GPU batch size from 320 seconds to 600 seconds improves WER by 3-4\% relative (C3), supporting our hypothesis that incorporating more positive and negative examples improves learned representations during the CSL pre-training. 
Gradient accumulation, which comes at no extra GPU memory cost, improves learning stability as each update observes more speech units. It provides a 3-6\% relative WER improvement (C7) and points to a potential future direction of adding a memory bank to hold extra positive and negative samples. Label-Aware Batching (LAB) provides a small but consistent boost in performance (C9) and provides the best performance both for the CSL and CE-PL systems when combined with all other improvements (C9 and B3).



\subsubsection{Effect of positives and negatives sampling policy}
In line with findings in self-supervised speech representation learning~\cite{w2v2, cpc}, using negative samples from the same utterance offers better WER compared to a random sampling of negative examples across utterances. The model learns to marginalize the speaker identity information while focusing on the speech content, which is preferable for an ASR downstream task but not necessarily for other speech applications. Using eight samples for both negative and positive comparisons was enough to reach the best performance. Limiting positives samples to be from the same or different utterances did not make a significant difference.


\subsubsection{Effect of best policy on CE-PL}
The last row of Table \ref{tab1} presents the best CSL results by combining all the improvements above. Applying the same improvements to cross-entropy pseudo-labeling (CE-PL) offers about a 15\% and a 12\% relative WER reduction (B3) compared to the original policy (B1).


\subsection{Combining CE-PL and CSL}
Table \ref{tab2} shows the experiments to test potential complementarity between CE pseudo-labeling (CE-PL) loss and the contrastive loss.  Using the setup of (C3) from Table \ref{tab1}, the two loss functions are applied to mini-batches using shown percentages in Table \ref{tab2}. Applying the CE-PL to some mini-batches does not improve the learned representation for the downstream ASR task than the CSL case.



\begin{table}
\centering
\setlength\tabcolsep{6.0pt}
 \begin{tabular}{cccc} 
\toprule
  & \textbf{CE-PL} &\textbf{CSL} & \textbf{WERR} \\ 
  \midrule\midrule
British English Videos & 32.0 & 29.4 & 8.1 \\ 
General English Videos & 37.2 & 32.8  & 11.8\\ 
Message Dictation & 21.6 & 17.8  & 17.3\\ 
Long-form Conversation & 26.0 & 22.0  & 15.4 \\ 

\bottomrule
\end{tabular}
\caption{Relative WERR (WER reduction) of using CSL and CE-PL for out of domain conditions}
\label{tab3}
\end{table}
\subsection{Out of Domain Generalization of CSL}

Table \ref{tab3} examines the generalization performance to out-of-domain test sets for British English. We compare two models pre-trained using CE-PL and CSL then fine-tuned in a low-resource condition using only 10hr labels. Although CSL offers about an 8\% relative improvement under matching training and testing conditions, it brought more significant improvements 11-17\% when tested in out-of-domain conditions. The CSL pre-training discovers more stable representations compared to CE-PL, which confirms our initial intuition.
\begin{table}
\centering
\setlength\tabcolsep{6.0pt}
 \begin{tabular}{cccc} 
\toprule
  & \textbf{CE-PL} &\textbf{CSL} & \textbf{WERR} \\ 
  \midrule\midrule
Generation 1 & 37.5 & 34.8 & 7.2 \\ 
Generation 2 & 33.1 & 30.9 & 6.6 \\ 
Generation 3 & 31.5 & 29.4 & 6.7 \\ 
\bottomrule
\end{tabular}
\caption{Effect of three iterations of data re-labeling on CSL pre-training.}
\label{tab4}
\end{table}
\subsection{Effect of iterative labeling}
Using multiple rounds of pseudo-labeling for model pre-training was found very effective for the classical GMM/HMM systems in the mid 2000s~\cite{wessel05} and more recently for neural systems~\cite{park2020improved, xu2020iterative, singh2020large}. Table \ref{tab4} examines if the gain observed using CSL diminishes under an iterative labeling scheme due to the improved baseline CE-PL performance. Each generation starts by generating new teacher pseudo-labels for the unlabeled videos using the previous generation's model. A CE fine-tuning step is performed for all experiments reported in Table \ref{tab4}. Both CE-PL and CSL improve using iterative re-labeling. The gains from CSL pre-training are complementary to iterative re-labeling.

\begin{table}
\centering
\setlength\tabcolsep{6.0pt}
 \begin{tabular}{cccc} 
\toprule
  & \textbf{British English (WER)} \\ 
  \midrule\midrule
  Supervised 1hr & 80.5 \\
  \midrule\midrule
CE-PL & 53.1 \\ 
CSL & 42.8 \\ 
CSL Generation 2 & 32.3 \\ 
\midrule\midrule
  Supervised 650 Hours & 23.1 \\
\bottomrule
\end{tabular}
\caption{Ultra low-resource condition using 1hr of labels.}
\label{tab5}
\vspace{-3mm}
\end{table}

\subsection{Ultra low-resource condition with 1hr of labels}

To confirm CSL pre-training resilience to teacher pseudo-labeling errors, we test CSL and CE-PL under an ultra low-resource condition using only 1hr of labeled data for teacher training. Table \ref{tab5} shows the WER on British English with about 19\% relative improvement achieved by CSL compared to CE-PL, which suffers from teacher label noise. Using 1hr of labels and two iterations of CSL, we can close about 80\% of the performance gap with 650hr of supervised data in the challenging domain of social media video transcription. 

\section{Conclusion}

Inspired by the impressive performance of self-supervised representation learning in computer vision and speech, we introduced Contrastive Semi-supervised Learning (CSL), which applies a contrastive loss for semi-supervised ASR pre-training. For the challenging task of social media video transcription, using 75,000hr of unlabeled videos and only 10hr of labeled data, CSL offers an 8\% relative WER reduction compared to the strongest Cross-Entropy pseudo-labeling (CE-PL) baseline. CSL pre-training proved more resilient under out-of-domain conditions and even using 1hr of labeled data.


\bibliographystyle{IEEEbib}
\bibliography{refs}

\end{document}